\documentclass{article}

\usepackage[preprint]{corl_2021} 
\usepackage[utf8]{inputenc} 
\usepackage[T1]{fontenc}    
\usepackage{hyperref}       
\usepackage{url}            
\usepackage{booktabs}       
\usepackage{amsfonts}       
\usepackage{nicefrac}       
\usepackage{microtype}      
\usepackage{graphicx} 
\usepackage{subfigure}
\usepackage{hyperref}
\usepackage{bm}
\usepackage{algorithm}
\usepackage{algorithmic}
\usepackage{amsmath}
\newtheorem{theorem}{Theorem}

\title{Planning with Exploration: Addressing Dynamics Bottleneck in Model-based Reinforcement Learning}

\author{ Xiyao Wang,    Junge Zhang \thanks{Junge Zhang is corresponding author.} ,   Wenzhen Huang,    Qiyue Yin
\\
  Center for Research on Intelligent System and Engineering \\
 Institute of Automation, CAS  \\
University of Chinese Academy of Sciences\\
  \texttt{xiyaowang96@gmail.com, jgzhang@nlpr.ia.ac.cn, } \\
  \texttt{huangwenzhen2014@ia.ac.cn, qyyin@nlpr.ia.ac.cn} \\
}

\begin{document}
\maketitle


\begin{abstract}
Model-based reinforcement learning (MBRL) is believed to have higher sample efficiency compared with model-free reinforcement learning (MFRL). 
However, MBRL is plagued by dynamics bottleneck dilemma. 
Dynamics bottleneck dilemma is the phenomenon that the performance of the algorithm falls into the local optimum instead of increasing when the interaction step with the environment increases, which means more data can not bring better performance. 
In this paper, we find that the trajectory reward estimation error is the main reason that causes dynamics bottleneck dilemma through theoretical analysis. 
We give an upper bound of the trajectory reward estimation error and point out that increasing the agent's exploration ability is the key to reduce trajectory reward estimation error, thereby alleviating dynamics bottleneck dilemma.
Motivated by this, a model-based control method combined with exploration named MOdel-based Progressive Entropy-based Exploration (MOPE2) is proposed. 
We conduct experiments on several complex continuous control benchmark tasks. The results verify that MOPE2 can effectively alleviate dynamics bottleneck dilemma and have higher sample efficiency than previous MBRL and MFRL algorithms.
\end{abstract}



\section{Introduction}
	
Reinforcement learning (RL) is one of the paradigms and methodologies of machine learning, which is used to describe and solve the problem that agents maximize returns or achieve specific goals through learning policies from the process of interacting with the environment. In recent years, reinforcement learning has achieved considerable success in many complex tasks, such as the Arcade-Learning environment \cite{bellemare2013arcade} and Alpha Go \cite{silver2016mastering}. 
Depending on whether modeling the environment, RL can be divided into two categories: model-free reinforcement learning (MFRL) and model-based reinforcement learning (MBRL).
MFRL algorithms \cite{{fujimoto2018addressing}, {haarnoja2018soft}, {lillicrap2015continuous}, {mnih2016asynchronous}} can achieve promising performance in many benchmark environments (MuJoCo \cite{6386109}, Atari \cite{1606.01540}, etc.). But they need massive samples to train the policy and it may be very expensive or even impossible to obtain so many samples in some real-world tasks such as autonomous driving and industrial robot \cite{dulac2019challenges}. 
 
MBRL is considered to be a solution to improve sample efficiency for it can improve the policy based on the samples generated by the learned dynamics model \cite{janner2019trust} instead of interacting with the real environment.
The research of MBRL progresses rapidly in recent years and many MBRL algorithms \cite{{chua2018deep},{clavera2018model}, {janner2019trust}, {kurutach2018model},{shen2020model}, {wang2019exploring}} significantly outperform MFRL algorithms with high sample efficiency in continuous control tasks and video games.
Meanwhile, in many benchmark environments, some MBRL algorithms such as probabilistic ensembles with trajectory sampling (PETS) \cite{chua2018deep} surpass the performance of MFRL algorithms in limited interaction steps with the environment. 
However, when the interaction steps increase, the performance of MBRL methods often falls into a suboptimal performance and stops increasing which means more samples do not result in better performance. 
This problem is named dynamics bottleneck dilemma and it is one of the main reason for the asymptotic performance gap between MBRL and MFRL methods   
which limits the application of MBRL methods in real-world tasks \cite{wang2019benchmarking}.

In this paper, we study the reason of dynamics bottleneck dilemma and how to alleviate dynamics bottleneck dilemma.
We first give an upper bound of the rollout trajectory's total reward gap between the learned dynamics model and the real environment.
By analyzing this upper bound, we find that trajectory reward estimation error caused by model error and the reward estimation gap will result in a suboptimal solution. 
And this is the main reason that causes dynamics bottleneck dilemma.
Then we propose that increasing the exploration ability of the agent can bring diverse samples and reduce trajectory reward estimation error, thereby alleviating dynamics bottleneck dilemma.
This inspires us to design MOdel-based Progressive Entropy-based Exploration (MOPE2), a novel model-based control method combined with exploration.
MOPE2 uses the entropy of the next state distribution predicted by the learned dynamics model as the intrinsic reward to increase the agent's exploration ability. 
Moreover, to solve the failure of typical exploration methods on MBRL, we propose the progressive exploration mechanism to gradually improve the agent's exploration ability.
To our knowledge, MOPE2 is the first algorithm proposed to solve dynamics bottleneck dilemma.
Experiments on several challenging continuous control benchmark tasks show that MOPE2 is a high sample efficiency method which can effectively alleviate dynamics bottleneck dilemma and greatly reduce the asymptotic performance gap between MBRL and MFRL methods 

\section{Related Work}
\subsection{Model-based Reinforcement Learning}
\label{sec2.1}
Model-based reinforcement learning (MBRL) has always been a research hotspot in reinforcement learning, and has made some remarkable progress in recent years. 
MBRL mainly focuses on two problems: how to learn a better model and how to use the model to obtain a better policy.
To learn a better model, Probabilistic Inference for Learning Control (PILCO) \cite{deisenroth2011pilco} and Guided Policy Search (GPS) \cite{montgomery2016guided} used a Gaussian process to model the environment.
Kurutach et al.  \cite{kurutach2018model} proposed to use dynamics model ensemble and Chua et al. proposed PETS \cite{chua2018deep} which extends the ensemble learning to probabilistic neural network to capture both aleatoric uncertainty and epistemic uncertainty. Shen et al. \cite{shen2020model} introduced adaptation augmented model-based policy optimization (AMPO) to reduce the potential distribution mismatch between simulated data and real data. 
To obtain a better policy, Luo et al. proposed Stochastic Lower Bounds Optimization (SLBO) \cite{luo2021algorithmic} which provides a framework that guarantees monotonic increase in policy's performance.
 Short \cite{janner2019trust} and bidirectional \cite{lai2020bidirectional} rollout schemes are designed to obtain more accurate samples training the policy. Meta-learning is used to improve the robustness of the policy to different environments in model-based meta-policy-optimzation (MB-MPO) \cite{clavera2018model}. The model-based control method \cite{{chua2018deep}, {POLO}} is very popular recently. Model-based control (MBC) method is a kind of MBRL method which uses model predictive control (MPC) as the policy. It is more robust to the task horizon and easier to calculate than the traditional model-based actor-critic methods. The method we proposed in this paper also belongs to MBC method.

\subsection{Exploration in Reinforcement Learning}
\label{sec2.2}

Exploration has been studied for quite a long time in reinforcement learning. 
One approach for better exploration uses intrinsic rewards as exploration bonus, which compose of two terms:
$r = r^{ext} + \beta r^{int}$,
where $r^{ext}$ is the extrinsic reward provided by environment and ${r^{int}}$ is the intrinsic reward computed by agent. $\beta$ is the temperature parameter. There are a lot of works on how to calculate the intrinsic reward recently. Pathak et al.  \cite{pathak2017curiosity} introduced an Intrinsic Curiosity Module to encourage the agent to explore the actions that it may be interested in. Pseudo-counts \cite{{bellemare2016unifying},{ostrovski2017count}} is another way to compute the intrinsic reward based on the estimated count of different states given by a density model. Pseudo-counts hope the agent visit the states with a low count as much as possible. Beside these, Burda et al. \cite{burda2018exploration} proposed Random Network Distillation (RND) which treats the intrinsic reward as the state prediction error of a random network. The agent should explore the states with high prediction errors. 
The entropy-based exploration \cite{{haarnoja2017reinforcement}, {haarnoja2018soft}} is proposed recently which the entropy of the action distribution generated by the policy network as the intrinsic reward.
However, typical exploration methods cannot be effectively used in MBRL for they do not consider the model.
Exploration in MBRL is not only carried out in the real environment, but also in the model. If the model is not considered, it will lead to wrong exploration. 
In subsequent sections we will discuss how to design effective exploration methods in MBRL.


\section{Background}

\subsection{Preliminary}

In this section, we will define some symbols used in reinforcement learning. We consider infinite-horizon Markov decision processes (MDP) defined by the tuple $(\mathcal{S}, \mathcal{A}, r, f, \gamma)$. We define action space as $\mathcal{A}$ and state space as $\mathcal{S}$. For the current state $s_t\in \mathcal{S}$ and action $a_t \in \mathcal{A}$, $r$ denote the reward $r(s_t, a_t)$ for a given state-action pair $(s_t, a_t)$ and $f(s_{t+1}|s_t,a_t)$ denote the transition function where $s_{t+1}$ is the next state.  $\gamma \in [0,1)$ is the discount factor. Agents should find the optimal policy $\pi$ by maximizing the total reward:

\begin{equation}\label{eq3_1}
\begin{split}
\pi= \mathop{\arg\max}\limits_{\pi}\mathbb{E}_{\pi,f}[\sum_{t=0}^{\infty}\gamma^{t}r(s_t,a_t)]
\end{split}
\end{equation}

where $a_t \sim \pi(a_t|s_t), s_{t+1} \sim f(s_{t+1}|s_t,a_t)$. In MBRL, we denote the learned dynamics model as $\hat{f}(s_{t+1}|s_t,a_t)$. We also define $r_{m}(s_t, a_t)$ and $r_{e}(s_t, a_t)$ as the reward of $(s_t, a_t)$ given by the learned dynamics model and the real environment respectively.

\subsection{Dynamics Bottleneck Dilemma}

\begin{figure}[h]
\centering
\includegraphics[width=14cm,height=3.2cm]{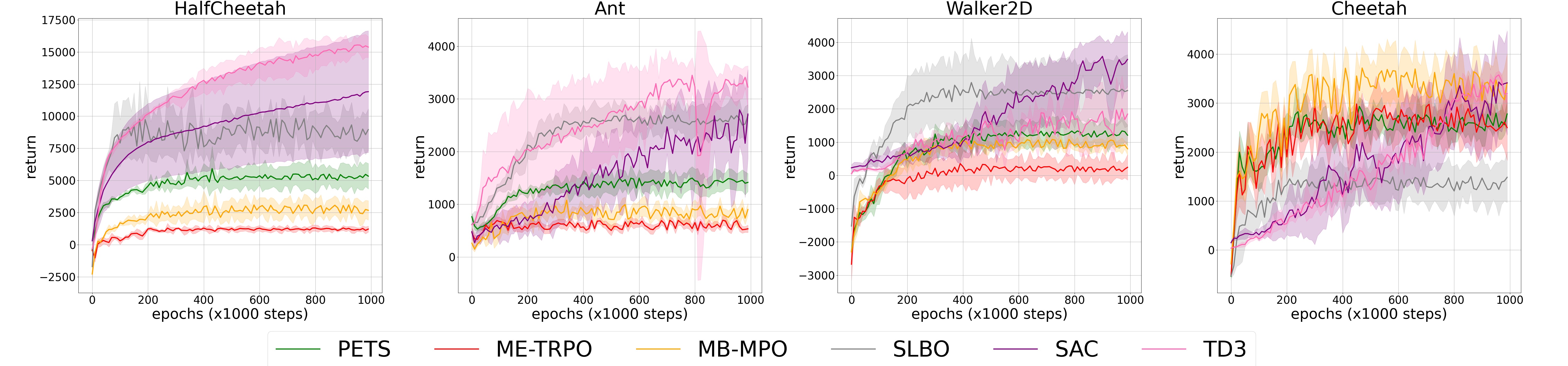}
\setlength{\abovecaptionskip}{-10pt}\caption{Performance curve for four MBRL algorithms and two MFRL algorithms trained for 1 million interaction steps.}
\label{Fig1}
\end{figure}

Dynamics bottleneck dilemma is an important problem that plagues model-based reinforcement learning. This concept is first proposed in \cite{wang2019benchmarking} by Wang et al.. 
As shown in Figure \ref{Fig1}, the performance of MBRL methods is converged when the number of interactive steps reaches 400k in the four tasks, while the performance of MFRL methods continues to increase and finally surpasses MBRL methods. 
Therefore, although MBRL has higher sample efficiency than MFRL, its asymptotic performance often lags behind MFRL a lot \footnote{Performance in this paper represents the cumulative reward obtained after the policy interacts with the evaluation environment for 1000 steps}.

\section{ Theoretical Analysis of Dynamics Bottleneck Dilemma}
\label{sec4}

In this section, we will theoretically analyze what causes dynamics bottleneck dilemma in MBRL and provide solutions to alleviate it.
In MBRL, to improve the sample efficiency, the rollout samples generated by the learned dynamics model are used to train the policy. 
So the update and iteration of the policy completely depends on the learned dynamics model. 
Under the learned dynamics model, the optimization problem in Eq. \ref{eq3_1} can be written as follows:

\begin{equation}\label{eq4_1}
\begin{split}
\pi= \mathop{\arg\max}\limits_{\pi}\mathbb{E}_{\pi,\hat{f}}[\sum_{t=0}^{H}\gamma^{t}r(s_t,a_t)] = \mathop{\arg\max}\limits_{\pi}\mathbb{E}_{\pi}[\underbrace{r_{m}(s_0, a_0) +\sum_{\tau = 1}^{H-1}\gamma^{\tau}r_{m}(\hat{f}(s_{\tau}|s_{\tau-1},a_{\tau-1}), \pi(s_{\tau}))]}_{\text{total  reward  of  trajectory}}
\end{split}
\end{equation}

where $H$ is the rollout length. MBRL tries to find the optimal policy using the rollout trajectory's total reward with the rollout length $H$. 
So the accuracy of calculating the trajectory's total reward is very important.

\begin{figure}[h]
\centering
\subfigure{
\includegraphics[width=13cm,height=7cm]{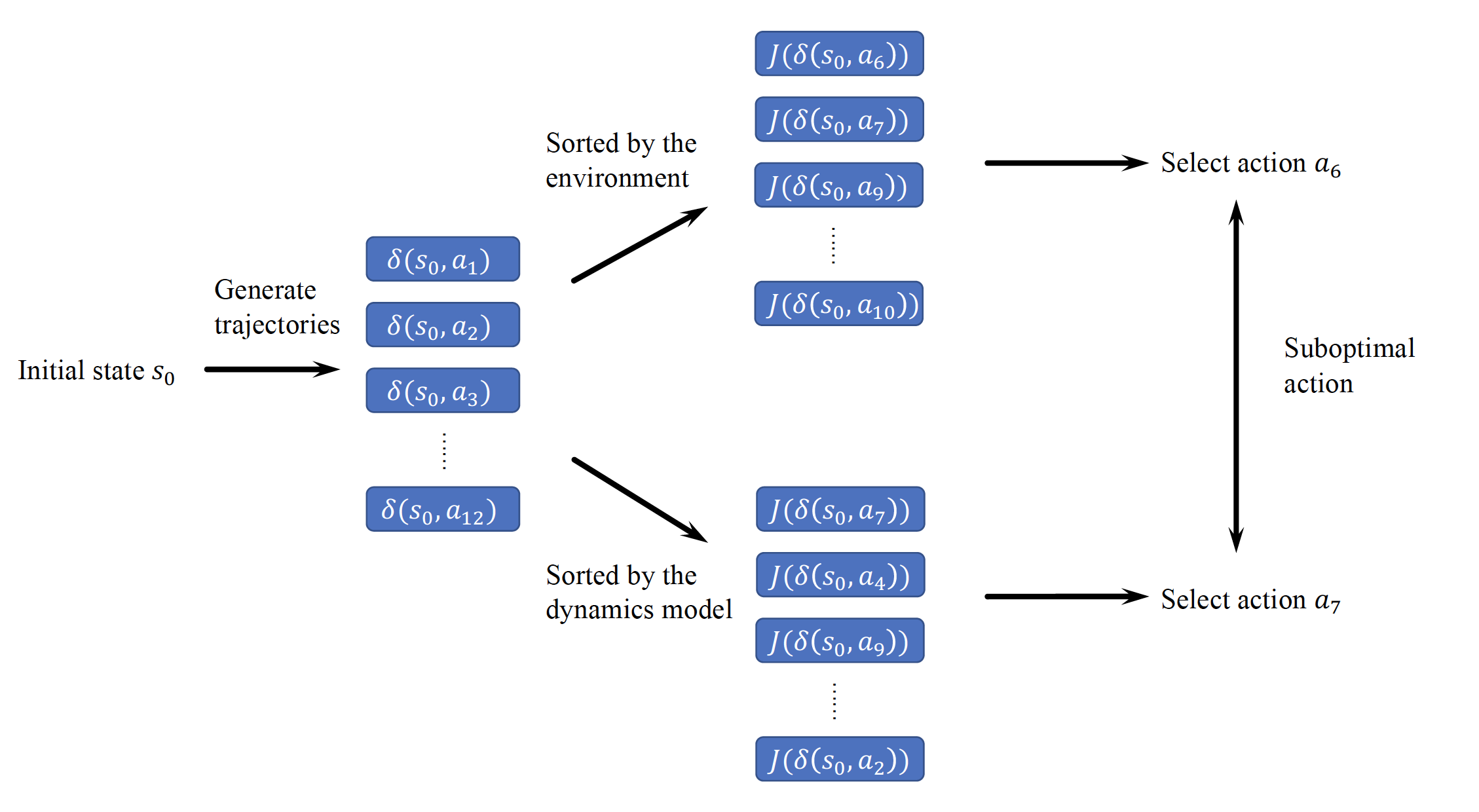}}
\caption{Example of suboptimal policy caused by TREE error in model-based control method. We assume the initial state as $s_0$ and generate 12 trajectories from $s_0$, namely $\delta(s_0,a_1)$ , $\delta(s_0,a_2)$...$\delta(s_0,a_{12})$. In the process of trajectory evaluation, due to the TREE error, the trajectories sorted by the learned dynamics model is inconsistent with the trajectories sorted by the real environment, which causes the agent to execute the suboptimal action $a_7$ instead of the best action $a_6$.}
\label{Fig2}
\end{figure}

We define the trajectory's total reward as  $J$.
In an ideal situation, the trajectory's total reward $J_{model}$ calculated by the reward function of the learned dynamics model is the same as the trajectory's total reward $J_{env}$ given by the real environment, then the agent can find the optimal policy using Eq. \ref{eq4_1}.
However, this is impossible to achieve this situation. 
In MBRL, there are two errors: the model error and the reward estimation gap. 
The model error refers to the the potential distribution mismatch between generated data and real data \cite{shen2020model} and the reward estimation gap refers to the difference between the reward estimated by the learned dynamics model's reward function and the reward given by the real environment.
Due to these two errors, there will always be a gap between $J_{model}$ and $J_{env}$. We name this gap Trajectory REward Estimation (TREE) error.
TREE error will cause the agent to incorrectly evaluate the value of the trajectory and fall into a suboptimal policy.
In Figure \ref{Fig2}, we provide an example of how TREE error lead to the suboptimal policy. 

Moreover, after the learned dynamics model is trained to convergence, the policy will also tend to stabilize. If the policy falls into a suboptimal solution, then the samples obtained by interacting with the environment will be suboptimal samples, which cannot bring substantial improvements to the model and the policy, causing the algorithm to fall into dynamics bottleneck dilemma.
From the above analysis, it can be concluded that the key to alleviate dynamics bottleneck dilemma is to narrow TREE error. 
Now we give our upper bound of TREE error:

\begin{theorem}
\label{th1}
For a state-action pair $(s_t, a_t)$, we define $\epsilon_r(s_t, a_t)$ as the reward gap between the reward given by the real environment and the learned dynamics model. And $\epsilon_{r_{max}}$ represents the maximum reward gap of all state-action pairs defined on $\mathcal{S}$ and $\mathcal{A}$. Denote the model error by $\epsilon_m = max_t[D_{TV}(p(s_{t+1}|s_t,a_t)||\hat{p}(s_{t+1}|s_t,a_t))]$. Given the trajectory $\delta(s_0, a_0)$ starting from $(s_0, a_0)$, the TREE error can be bounded as:
\begin{equation}\label{eq4_2}
\begin{split}
Error_{TREE} = J_{env}[\delta(s_0, a_0)]-J_{model}[\delta(s_0, a_0)] \leq \underbrace{\frac{(1-\gamma^{H})\epsilon_{r_{max}}}{1-\gamma}}_{\text{reward  estimation  gap}} + 
\underbrace{\frac{2r_{max}(\gamma-\gamma^{H})\epsilon_m}{1-\gamma}}_{\text{model  error}}
\end{split}
\end{equation}
\end{theorem}

\newenvironment{proof}{{\noindent\it Proof.}\quad}{\hfill $\square$\par}
\begin{proof}
See Appendix A1
\end{proof}

where $\gamma$ is the discount rate, $H$ is the rollout length and $r_{max}$ is the maximum reward of all state-action pairs defined on $\mathcal{S}$ and $\mathcal{A}$. 
From Theorem \ref{th1} we can find that as long as the model error and reward estimation gap are reduced, the TREE error can be reduced, thereby alleviating dynamics bottleneck dilemma. 
To reduce the model error, we can design better model learning or usage methods and a lot of previous work \cite{{chua2018deep}, {janner2019trust},{shen2020model},{lai2020bidirectional},{pan2020trust}} has focused on it.
However, how to reduce the reward estimation gap still remains an opening problem which is we focus on in this paper. 

To reduce the reward estimation gap, one of the main approachs is to obtain more diverse samples. 
Because the root cause of the reward estimation gap is that the replay buffer only has a part of samples in the sample space. This makes the reward function of the learned dynamics model cannot correctly estimate the reward for unseen samples.
So the diversity of samples is necessary.
The traditional RL algorithms obtain more diverse samples by increasing the exploration ability of the agent.
As introduced in Sec. \ref{sec2.2}, an intrinsic reward is added when calculating the reward of samples.
However, it is hard to directly use traditional exploration methods in MBRL. 
Because in MBRL, exploration must be carried out not only in the real environment, but also in the learned dynamics model. 
In the early stage of model learning, the dynamics model is usually inaccurate, which will invalidate the exploration.
Therefore, in order to increase the exploration ability of the agent in MBRL and alleviate dynamics bottleneck dilemma, we propose a novel model-based control method named MOdel-based Progressive Entropy-based Exploration (MOPE2) which is detailed in next section.

\section{Model-based Progressive Entropy-based Exploration}
\label{sec5}
In this section, we introduce our model-based control method: MOdel-based Progressive Entropy-based Exploration (MOPE2). In Sec. \ref{5.1} and \ref{5.2}, we will describe two important parts of MOPE2, i.e. the intrinsic reward and progressive exploration mechanism, respectively. Implementation details of MOPE2 will be introduced in Sec. \ref{5.3} .
\subsection{Intrinsic Reward}
\label{5.1}
As we have introduced in Sec. \ref{sec2.2}, traditional reinforcement learning methods improve the agent's exploration ability by adding an intrinsic reward to the total reward.
But using typical bonus-based exploration methods on MBRL cannot take advantage of the learned dynamics model.
In MOPE2, we design a new entropy-based intrinsic reward.
We propose to use the entropy of the next state distribution predicted by the learned dynamics model as the intrinsic reward. 
Given a state-action pair $(s,a)$, the probabilistic neural network ensemble is used as the learned dynamics model and outputs a  multivariate Gaussian distribution as the predicted state distribution.
The entropy of predicted state distribution $\hat{p}(\cdot|s_t, a_t)$ can be utilized as an estimate of epistemic uncertainty and aleatoric uncertainty in the prediction \cite{Abdar_2021}, so a state-action pair with high predicted state distribution entropy means a high probability of generating diverse samples.
The intrinsic reward can be calculated as follows:

\begin{equation}\label{eq5.1}
r^{int} = \mathcal{H}(\hat{p}(\cdot|s,a)) = \frac{k}{2}(ln2\pi+1)+\frac{1}{2}ln|\Sigma_{\hat{p}(\cdot|s,a)}|
\end{equation}

where $k$ is the dimension of the state and $\Sigma_{\hat{p}(\cdot|s,a)}$ is the covariance matrix of the predicted state distribution. Then the total reward of  the trajectory with length $H$ is:

\begin{equation}\label{eq5.2}
J(\delta)=\sum_{t=0}^{H-1}\gamma^{t}[r_m(s_t,a_t)+\beta(\frac{k}{2}(ln2\pi+1)+\frac{1}{2}ln|\Sigma_{\hat{p}(\cdot|s_t,a_t)}|)]
\end{equation}

where $s_t \sim \hat{p}(\cdot|s_{t-1},a_{t-1})$ and $\beta$ is the temperature parameter.
This intrinsic reward can drive the agent to explore regions with high epistemic uncertainty and aleatoric uncertainty during planning, thereby bringing more diverse samples and improving the accuracy of reward estimation.

\subsection{Progressive Exploration Mechanism}
\label{5.2}
In Eq. \ref{eq5.2}, the intrinsic reward is added to the extrinsic reward $r_m$ directly with a fixed temperature parameter. But as we have discussed in Sec. \ref{sec4}, the prediction of the learned dynamics model is usually inaccurate at the early stage of model learning. This will make the agent calculate the entropy incorrectly for the entropy is directly related to the prediction of the learned dynamics model. If the entropy is directly added to the reward, the entropy will mislead the exploration and result in poor performance of the algorithm. 

To solve this problem, we propose progressive exploration mechanism which makes the temperature parameter increase linearly with the model learning. Define the interaction epoch with the environment as $e$ and each epoch contains 1000 steps. The learned dynamics model is trained after each interaction epoch based on the samples obtained from the interaction epoch. 
Before each interaction epoch, the temperature parameter $\beta$ is calculated as follows:

\begin{equation}\label{eq5.3}
\beta = f(e) = min(max(\beta_{min}+\frac{e-e_{min}}{e_{max}-e_{min}}, \beta_{min}), \beta_{max})
\end{equation}

where $\beta_{max}$ and $\beta_{min}$ are the upper threshold and lower threshold of $\beta$, $e_{max}$ and $e_{min}$ are the upper threshold and lower threshold of $e$. We name Eq. \ref{eq5.3} progressive exploration mechanism. When the interactive epoch is lower than $_{min}$, $\beta$ is set to 0 and the agent will not explore for the learned dynamics model is not accurate enough. 
After the interaction epoch exceeds the lower threshold $e_{min}$, $\beta$ increases linearly from the lower threshold $\beta_{min}$ with the increase of the interaction epoch $e$. This is a progressive exploration process. As the learned dynamics model becomes more and more accurate, the agent's exploration ability continues to increase, and finally reaches the upper threshold $\beta_{max}$.
We demonstrate the effectiveness and importance of our mechanism through experiments in appendix.

\subsection{Implementation}
\label{5.3}
In this section, we will introduce the implementation details of MOPE2 and the pseudo code is in Algorithm \ref{Alg1}.

\begin{algorithm}
\caption{MOPE2}
\label{Alg1}
\begin{algorithmic}[1]
\STATE Initialize dynamics model parameters $\theta$, reward function parameters $\phi$, dataset $\mathbb{D}$, planning horizon $\emph{H}$, update rate $\alpha$ \
\FOR{each interaction epoch $e$} 
\FOR{ $\emph{i}^{th}$ interaction step} 
\STATE Initialize action distribution. $\mu = \mu_{0}$ , $\Sigma = \sigma_{0}^{2} \emph{\textbf{I}}$ \ 
\WHILE{not convergence}
\STATE Generate $K$ trajectories using dynamics model and action distribution \
\STATE Compute temperature parameter $\beta$ using Eq. \ref{eq5.3} and trajectories' total reward using reward function and Eq. \ref{eq5.2}
\STATE Fit current action distribution $\mu^{'}$ , $\Sigma^{'}$ using the first action of the top $k$ largest total reward trajectories\
\STATE Update action distribution by $\mu = (1-\alpha)\mu +\alpha \mu^{'} , \Sigma = (1-\alpha)\Sigma +\alpha \Sigma^{'}$ \
\ENDWHILE
\STATE Execute the first action of the optimal trajectory \
\STATE Record outcome:$\mathbb{D}\gets \mathbb{D}\cup \{s_t, a_t, s_{t+1}, r_t\}$ \
\ENDFOR 
\STATE Update $\theta$ and $\phi$ using $\mathbb{D}$ \
\ENDFOR 
\end{algorithmic}
\end{algorithm}

\paragraph{The Learned Dynamics Model} In MOPE2, we use the probabilistic neural network ensemble as the learned dynamics model which is the same as PETS. For a single probabilistic neural network, we define our predictive model to output a Gaussian distribution with diagonal covariances parameterized by $\theta$ and conditioned on $s_t$ and $a_t$, i.e.: $\hat{p}(s_{t+1}|s_t,a_t) = \mathcal{N}(\mu_{\theta}(s_t,a_t),\Sigma_{\theta}(s_t,a_t))$. We train the model based on the samples obtained from interaction with the environment and the loss function is:

\begin{equation}\label{eq5.4}
loss = \sum_{n=1}^N [\mu_{\theta}(s_n,a_n)-s_{n+1}]^\top \Sigma_{\theta}^{-1}(s_n,a_n) [\mu_{\theta}(s_n,a_n)-s_{n+1}] + logdet\Sigma_{\theta}(s_n,a_n)
\end{equation}

Meanwhile, we use a neural network as the reward function which is trained using MSE loss.

\paragraph{Planning} We use MPC as our policy for its lower computational burden and more robust to task horizon than actor-critic methods  \cite{chua2018deep}. 
We initialize the action distribution as a multivariate Gaussian distribution with mean $\mu = \mu_0$ and covariance  $\Sigma = \sigma_{0}^{2} \emph{\textbf{I}}$ and sample $K$ action sequences from the action distribution with horizon $H$. 
Then we use the learned dynamics model to generate the state sequences based on the action sequences. For these $K$ trajectories, we calculate the temperature parameter $\beta$ using Eq. \ref{eq5.3} and Eq. \ref{eq5.2} is used to calculate the trajectories' total reward. We choose the first action of the top $k$ largest total reward trajectories to update the action distribution. The process above is repeated until the total reward of the best trajectory converges and the agent execute the first action of the optimal trajectory to interact with the environment.

\section{Experiments}

\begin{figure}[h]
\centering           
\includegraphics[width=14cm,height=3.2cm]{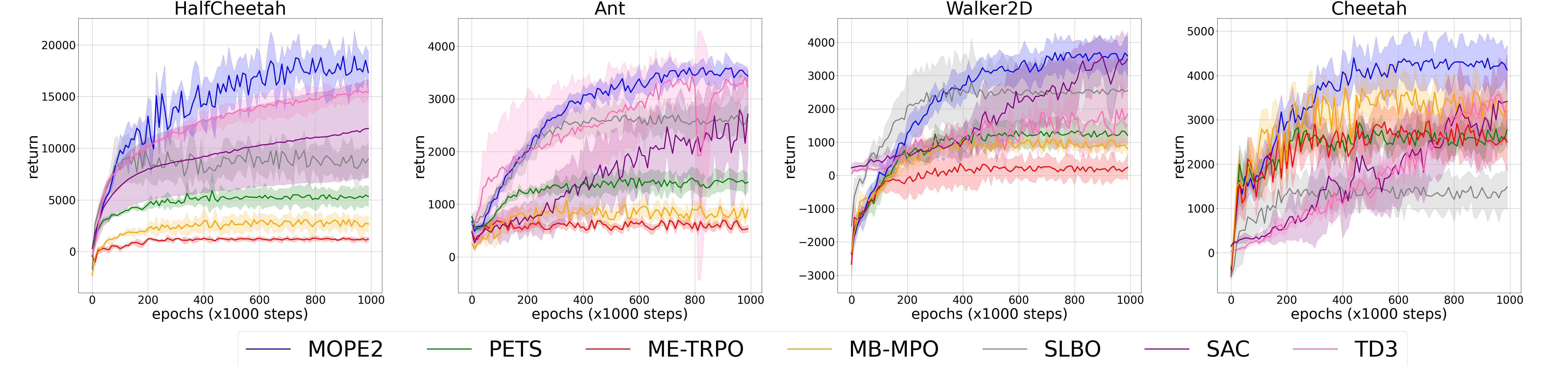}
\setlength{\abovecaptionskip}{-10pt}\caption{Performance curve for MOPE2, PETS, SAC and TD3. The solid curve represents the mean of eight trials and shaded areas indicate the standard deviation.}
\label{Fig4}
\end{figure}

In this section, we compare MOPE2 with four MBRL algorithms and two MFRL algorithms \footnote{Due to space limitations, more benchmark experiments and ablation experiments are in the appendix.}.
For the MBRL algorithms, we choose PETS \cite{chua2018deep} which is the state-of-the-art model-based control algorithm, SLBO \cite{luo2021algorithmic} which is one of the state-of-the-art MBRL algorithms, ME-TRPO  \cite{kurutach2018model} and MB-MPO \cite{clavera2018model}. 
For the MFRL baseline, we compare with two state-of-the-art MFRL algorithms SAC \cite{haarnoja2018soft} and TD3 \cite{fujimoto2018addressing}. 
Experiments are conducted on four complex continuous control tasks of the MuJoCo-v2 benchmark, HalfCheetah, Ant, Walker2D and Cheetah. Each algorithm runs eight random seeds and the interaction step is 1 million.
We set $\beta_{min}$, $\beta_{max}$, $e_{min}$ and $e_{max}$ in MOPE2 as 0, 1, 50, 300 respectively and model ensemble size is 4.
The performance curve is given in Figure \ref{Fig4}, and convergence performance is summarized in Table \ref{tab3}.

\begin{table}[h]
\setlength{\abovecaptionskip}{0.2cm}
\centering
\caption{Convergence performance of MOPE2, PETS, ME-TRPO, MB-MPO, SLBO, SAC and TD3. The results show the mean and standard deviation averaged over eight random seeds at last 5000 interaction steps.}
\begin{tabular}{ccccc}
\toprule 
&HalfCheetah&Ant&Walker2D&Cheetah \\
\midrule
MOPE2& $\mathbf{18469.6\pm3727.6}$ & $\mathbf{3440.8\pm277.3}$ & $\mathbf{3543.9\pm576.7}$ & $\mathbf{4387.6\pm580.8}$  \\
PETS& $5521.7\pm919.2$ &$1382.9\pm246.8$&$1197.9\pm975.8$ & $2772.7\pm966.8$ \\
ME-TRPO& $1084.6\pm 336.8$ &$679.8\pm 34.4$&$206.4\pm 130.2$ & $2333.4\pm 859.4$ \\
MB-MPO& $2142.0\pm 973.2$ &$814.4\pm 130.2$&$918.8\pm 152.2$ & $3284.2\pm 595.2$ \\
SLBO& $9619.4\pm 685.0$ &$2729.4\pm 93.4$&$2574.0\pm 87.8$ & $1348.2\pm 438.2$ \\
\midrule
SAC& $16105.4\pm 3849.2$&$\mathbf{4373.6\pm 159.8}$ &$\mathbf{4711.7\pm 613.6}$ &$\mathbf{4697.3\pm 725.4}$ \\
TD3& $\mathbf{18884.2\pm 1967.0}$&$3252.7\pm 186.2$ &$3979.5\pm 603.4$ &$3742.8\pm 632.4$ \\
\bottomrule
\end{tabular}
\label{tab3}
\end{table}

From the comparison in Figure \ref{Fig4} and Table \ref{tab3}, it can be seen that the performance in all four tasks has been greatly improved by MOPE2 compared with the previous model-based algorithms. 
And MOPE2 greatly reduces the asymptotic performance gap between MBRL and MFRL algorithms.
It is worth noting that MOPE2 performs three times better than the state-of-the-art model-based control method PETS. 
This proves the effectiveness of planning with exploration. 
And MOPE2 also has a high sample efficiency, only needs one-fifth to one-quarter interaction steps of the MFRL algorithm to achieve the same performance. 
Meanwhile, we find that the performance of other MBRL algorithms has converged when the interaction steps reach about 400k in these four tasks. 
While the performance of MOPE2 is still increasing at 500k interaction steps. 
In Table \ref{tab4} we summarize the number of interaction steps required for MOPE2 and the other four MBRL algorithms to achieve the convergence performance.
We can find that the number of interaction steps required for MEPO2 to achieve convergence performance is greatly increased compared with other MBRL algorithms, which means that more samples bring better performance, and dynamic bottleneck dilemma is greatly alleviated.

\begin{table}[h]
\setlength{\abovecaptionskip}{0.2cm}
\centering
\caption{The number of interaction steps required for MOPE2 and the other four MBRL algorithms to achieve the convergence performance in four tasks.}
\begin{tabular}{ccccc}
\toprule 
&HalfCheetah&Ant&Walker2D&Cheetah \\
\midrule
MOPE2& $780k$ & $700k$ & $750k$ & $590k$  \\
PETS& $300k$ &$320k$&$250k$ & $330k$ \\
ME-TRPO& $240k$ &$50k$&$200k$ & $280k$ \\
MB-MPO& $400k$ &$100k$&$360k$ & $300k$ \\
SLBO& $120k$ &$430k$&$400k$ & $250k$ \\
\bottomrule
\end{tabular}
\label{tab4}
\end{table}

\begin{figure}[h]
\centering
\subfigure{
\includegraphics[width=11cm,height=5cm]{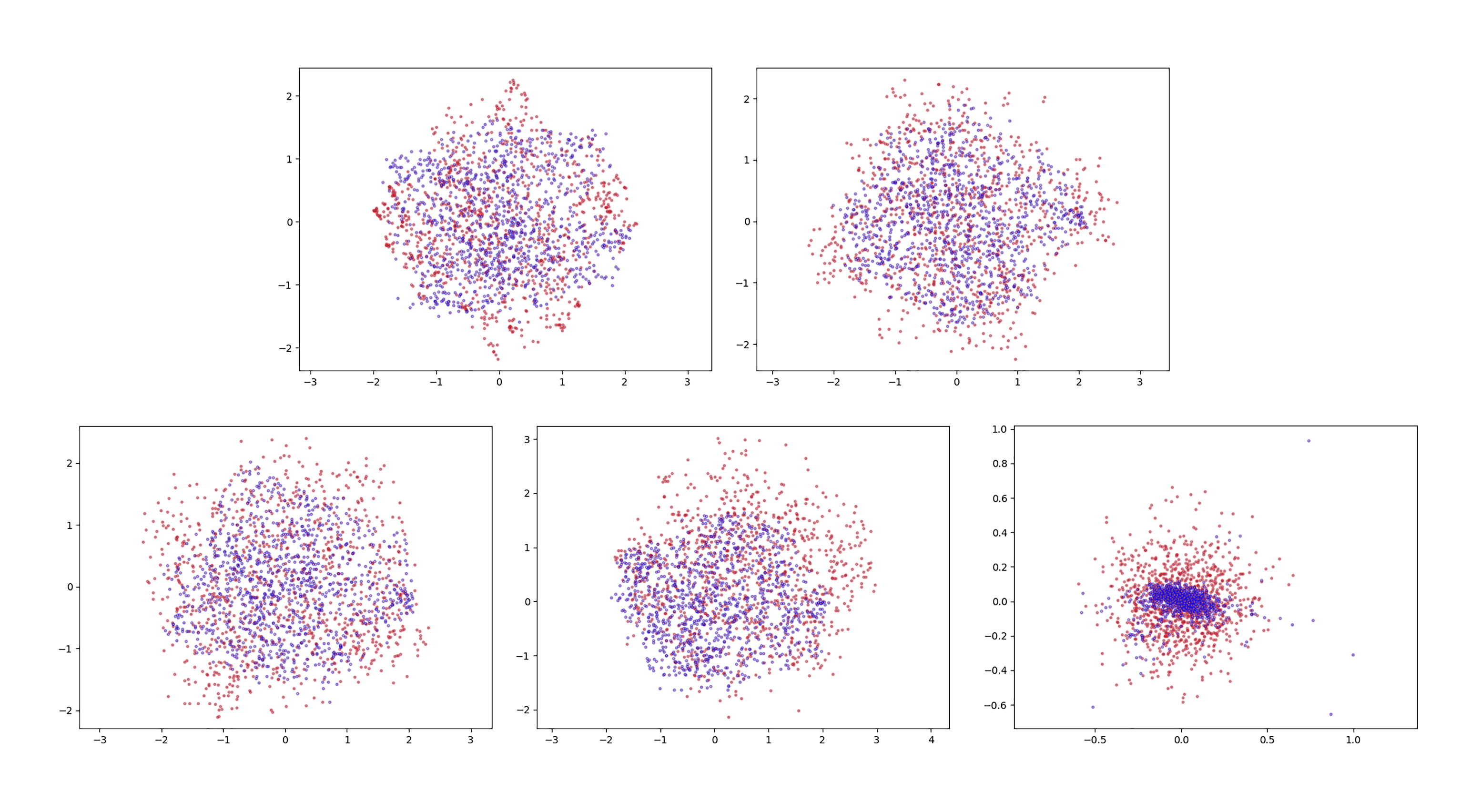}}
\caption{Action distribution of MOPE2 and PETS after PCA. Red dots represent MOPE2 and blue dots represent PETS. The interaction epochs from top left to bottom right are 0, 100, 200, 300 and 400 respectively. The horizontal axis and the vertical axis represent the most important and second most important dimensions of action.}
\label{Fig5}
\end{figure}

To prove that our exploration method improves sample diversity, we also analyze the distribution of actions. 
We select the actions executed by MEPO2 and PETS on the Ant task in 5 different interaction epochs (the epochs are 0, 100, 200, 300, and 400). Each epoch contains 1000 interaction steps which means 1000 actions.
We use PCA to retain the two most important dimensions of the actions and make a 2D visualization of the action distribution. 
It can be seen from the Figure \ref{Fig5} that the action distribution range of MECEM is significantly larger than that of PETS when the number of interaction epoch increases, indicating that MOPE2 effectively improves the sample diversity.

\section{Conclusions}
In this paper, we study an important problem that plagues MBRL called dynamics bottleneck dilemma. Through theoretical analysis, we find that the trajectory reward estimated gap between the dynamic model and the real environment is the main reason that causes the dynamics bottleneck dilemma, and increasing the diversity of samples by improving the agent's exploration ability can narrow this gap and alleviate the dynamics bottleneck dilemma. Inspired by this, we propose a novel model-based control method combined with exploration named MOdel-based Progressive Entropy-based Exploration (MOPE2). Experimental results show that MOPE2 can effectively alleviate dynamic bottlenecks and has high sample efficiency compared with the previoust MBRL and MFRL methods. To our knowledge, MOPE2 is the first model-based algorithm which is proposed  to solve the dynamics bottleneck dilemma. Future work may include designing more complex exploration methods in MBRL and using them in real-world tasks.


\bibliographystyle{plain}
\bibliography{ref}  

\begin{thebibliography}{28}
\providecommand{\natexlab}[1]{#1}
\providecommand{\url}[1]{\texttt{#1}}
\expandafter\ifx\csname urlstyle\endcsname\relax
  \providecommand{\doi}[1]{doi: #1}\else
  \providecommand{\doi}{doi: \begingroup \urlstyle{rm}\Url}\fi

\bibitem[Bellemare et~al.(2013)Bellemare, Naddaf, Veness, and
  Bowling]{bellemare2013arcade}
M.~G. Bellemare, Y.~Naddaf, J.~Veness, and M.~Bowling.
\newblock The arcade learning environment: An evaluation platform for general
  agents.
\newblock \emph{Journal of Artificial Intelligence Research}, 47:\penalty0
  253--279, 2013.

\bibitem[Silver et~al.(2016)Silver, Huang, Maddison, Guez, Sifre, Van
  Den~Driessche, Schrittwieser, Antonoglou, Panneershelvam, Lanctot,
  et~al.]{silver2016mastering}
D.~Silver, A.~Huang, C.~J. Maddison, A.~Guez, L.~Sifre, G.~Van Den~Driessche,
  J.~Schrittwieser, I.~Antonoglou, V.~Panneershelvam, M.~Lanctot, et~al.
\newblock Mastering the game of go with deep neural networks and tree search.
\newblock \emph{nature}, 529\penalty0 (7587):\penalty0 484, 2016.

\bibitem[Fujimoto et~al.(2018)Fujimoto, Van~Hoof, and
  Meger]{fujimoto2018addressing}
S.~Fujimoto, H.~Van~Hoof, and D.~Meger.
\newblock Addressing function approximation error in actor-critic methods.
\newblock \emph{arXiv preprint arXiv:1802.09477}, 2018.

\bibitem[Haarnoja et~al.(2018)Haarnoja, Zhou, Abbeel, and
  Levine]{haarnoja2018soft}
T.~Haarnoja, A.~Zhou, P.~Abbeel, and S.~Levine.
\newblock Soft actor-critic: Off-policy maximum entropy deep reinforcement
  learning with a stochastic actor.
\newblock \emph{arXiv preprint arXiv:1801.01290}, 2018.

\bibitem[Lillicrap et~al.(2015)Lillicrap, Hunt, Pritzel, Heess, Erez, Tassa,
  Silver, and Wierstra]{lillicrap2015continuous}
T.~P. Lillicrap, J.~J. Hunt, A.~Pritzel, N.~Heess, T.~Erez, Y.~Tassa,
  D.~Silver, and D.~Wierstra.
\newblock Continuous control with deep reinforcement learning.
\newblock \emph{arXiv preprint arXiv:1509.02971}, 2015.

\bibitem[Mnih et~al.(2016)Mnih, Badia, Mirza, Graves, Lillicrap, Harley,
  Silver, and Kavukcuoglu]{mnih2016asynchronous}
V.~Mnih, A.~P. Badia, M.~Mirza, A.~Graves, T.~Lillicrap, T.~Harley, D.~Silver,
  and K.~Kavukcuoglu.
\newblock Asynchronous methods for deep reinforcement learning.
\newblock In \emph{International conference on machine learning}, pages
  1928--1937, 2016.

\bibitem[{Todorov} et~al.(2012){Todorov}, {Erez}, and {Tassa}]{6386109}
E.~{Todorov}, T.~{Erez}, and Y.~{Tassa}.
\newblock Mujoco: A physics engine for model-based control.
\newblock In \emph{2012 IEEE/RSJ International Conference on Intelligent Robots
  and Systems}, pages 5026--5033, 2012.
\newblock \doi{10.1109/IROS.2012.6386109}.

\bibitem[Brockman et~al.(2016)Brockman, Cheung, Pettersson, Schneider,
  Schulman, Tang, and Zaremba]{1606.01540}
G.~Brockman, V.~Cheung, L.~Pettersson, J.~Schneider, J.~Schulman, J.~Tang, and
  W.~Zaremba.
\newblock Openai gym, 2016.

\bibitem[Dulac-Arnold et~al.(2019)Dulac-Arnold, Mankowitz, and
  Hester]{dulac2019challenges}
G.~Dulac-Arnold, D.~Mankowitz, and T.~Hester.
\newblock Challenges of real-world reinforcement learning.
\newblock \emph{arXiv preprint arXiv:1904.12901}, 2019.

\bibitem[Janner et~al.(2019)Janner, Fu, Zhang, and Levine]{janner2019trust}
M.~Janner, J.~Fu, M.~Zhang, and S.~Levine.
\newblock When to trust your model: Model-based policy optimization.
\newblock In \emph{Advances in Neural Information Processing Systems}, pages
  12519--12530, 2019.

\bibitem[Chua et~al.(2018)Chua, Calandra, McAllister, and Levine]{chua2018deep}
K.~Chua, R.~Calandra, R.~McAllister, and S.~Levine.
\newblock Deep reinforcement learning in a handful of trials using
  probabilistic dynamics models.
\newblock In \emph{Advances in Neural Information Processing Systems}, pages
  4754--4765, 2018.

\bibitem[Clavera et~al.(2018)Clavera, Rothfuss, Schulman, Fujita, Asfour, and
  Abbeel]{clavera2018model}
I.~Clavera, J.~Rothfuss, J.~Schulman, Y.~Fujita, T.~Asfour, and P.~Abbeel.
\newblock Model-based reinforcement learning via meta-policy optimization.
\newblock \emph{arXiv preprint arXiv:1809.05214}, 2018.

\bibitem[Kurutach et~al.(2018)Kurutach, Clavera, Duan, Tamar, and
  Abbeel]{kurutach2018model}
T.~Kurutach, I.~Clavera, Y.~Duan, A.~Tamar, and P.~Abbeel.
\newblock Model-ensemble trust-region policy optimization.
\newblock \emph{arXiv preprint arXiv:1802.10592}, 2018.

\bibitem[Shen et~al.(2020)Shen, Zhao, Zhang, and Yu]{shen2020model}
J.~Shen, H.~Zhao, W.~Zhang, and Y.~Yu.
\newblock Model-based policy optimization with unsupervised model adaptation.
\newblock \emph{Advances in Neural Information Processing Systems}, 33, 2020.

\bibitem[Wang and Ba(2019)]{wang2019exploring}
T.~Wang and J.~Ba.
\newblock Exploring model-based planning with policy networks.
\newblock \emph{arXiv preprint arXiv:1906.08649}, 2019.

\bibitem[Wang et~al.(2019)Wang, Bao, Clavera, Hoang, Wen, Langlois, Zhang,
  Zhang, Abbeel, and Ba]{wang2019benchmarking}
T.~Wang, X.~Bao, I.~Clavera, J.~Hoang, Y.~Wen, E.~Langlois, S.~Zhang, G.~Zhang,
  P.~Abbeel, and J.~Ba.
\newblock Benchmarking model-based reinforcement learning.
\newblock \emph{arXiv preprint arXiv:1907.02057}, 2019.

\bibitem[Deisenroth and Rasmussen(2011)]{deisenroth2011pilco}
M.~Deisenroth and C.~E. Rasmussen.
\newblock Pilco: A model-based and data-efficient approach to policy search.
\newblock In \emph{Proceedings of the 28th International Conference on machine
  learning (ICML-11)}, pages 465--472, 2011.

\bibitem[Montgomery and Levine(2016)]{montgomery2016guided}
W.~H. Montgomery and S.~Levine.
\newblock Guided policy search via approximate mirror descent.
\newblock In \emph{Advances in Neural Information Processing Systems}, pages
  4008--4016, 2016.

\bibitem[Luo et~al.(2021)Luo, Xu, Li, Tian, Darrell, and
  Ma]{luo2021algorithmic}
Y.~Luo, H.~Xu, Y.~Li, Y.~Tian, T.~Darrell, and T.~Ma.
\newblock Algorithmic framework for model-based deep reinforcement learning
  with theoretical guarantees, 2021.

\bibitem[Lai et~al.(2020)Lai, Shen, Zhang, and Yu]{lai2020bidirectional}
H.~Lai, J.~Shen, W.~Zhang, and Y.~Yu.
\newblock Bidirectional model-based policy optimization.
\newblock In \emph{International Conference on Machine Learning}, pages
  5618--5627. PMLR, 2020.

\bibitem[Lowrey et~al.(2019)Lowrey, Rajeswaran, Kakade, Todorov, and
  Mordatch]{POLO}
K.~Lowrey, A.~Rajeswaran, S.~Kakade, E.~Todorov, and I.~Mordatch.
\newblock {Plan Online, Learn Offline: Efficient Learning and Exploration via
  Model-Based Control}.
\newblock In \emph{{International Conference on Learning Representations
  (ICLR)}}, 2019.

\bibitem[Pathak et~al.(2017)Pathak, Agrawal, Efros, and
  Darrell]{pathak2017curiosity}
D.~Pathak, P.~Agrawal, A.~A. Efros, and T.~Darrell.
\newblock Curiosity-driven exploration by self-supervised prediction.
\newblock In \emph{International Conference on Machine Learning}, pages
  2778--2787. PMLR, 2017.

\bibitem[Bellemare et~al.(2016)Bellemare, Srinivasan, Ostrovski, Schaul,
  Saxton, and Munos]{bellemare2016unifying}
M.~G. Bellemare, S.~Srinivasan, G.~Ostrovski, T.~Schaul, D.~Saxton, and
  R.~Munos.
\newblock Unifying count-based exploration and intrinsic motivation.
\newblock \emph{arXiv preprint arXiv:1606.01868}, 2016.

\bibitem[Ostrovski et~al.(2017)Ostrovski, Bellemare, Oord, and
  Munos]{ostrovski2017count}
G.~Ostrovski, M.~G. Bellemare, A.~Oord, and R.~Munos.
\newblock Count-based exploration with neural density models.
\newblock In \emph{International conference on machine learning}, pages
  2721--2730. PMLR, 2017.

\bibitem[Burda et~al.(2018)Burda, Edwards, Storkey, and
  Klimov]{burda2018exploration}
Y.~Burda, H.~Edwards, A.~Storkey, and O.~Klimov.
\newblock Exploration by random network distillation.
\newblock \emph{arXiv preprint arXiv:1810.12894}, 2018.

\bibitem[Haarnoja et~al.(2017)Haarnoja, Tang, Abbeel, and
  Levine]{haarnoja2017reinforcement}
T.~Haarnoja, H.~Tang, P.~Abbeel, and S.~Levine.
\newblock Reinforcement learning with deep energy-based policies.
\newblock In \emph{International Conference on Machine Learning}, pages
  1352--1361. PMLR, 2017.

\bibitem[Pan et~al.(2020)Pan, He, Tu, and He]{pan2020trust}
F.~Pan, J.~He, D.~Tu, and Q.~He.
\newblock Trust the model when it is confident: Masked model-based
  actor-critic.
\newblock \emph{Advances in Neural Information Processing Systems}, 2020.

\bibitem[Abdar et~al.(2021)Abdar, Pourpanah, Hussain, Rezazadegan, Liu,
  Ghavamzadeh, Fieguth, Cao, Khosravi, Acharya, and et~al.]{Abdar_2021}
M.~Abdar, F.~Pourpanah, S.~Hussain, D.~Rezazadegan, L.~Liu, M.~Ghavamzadeh,
  P.~Fieguth, X.~Cao, A.~Khosravi, U.~R. Acharya, and et~al.
\newblock A review of uncertainty quantification in deep learning: Techniques,
  applications and challenges.
\newblock \emph{Information Fusion}, May 2021.
\newblock ISSN 1566-2535.
\newblock \doi{10.1016/j.inffus.2021.05.008}.
\newblock URL \url{http://dx.doi.org/10.1016/j.inffus.2021.05.008}.

\end{thebibliography}
\clearpage
\appendix
\section{Appendix}

\subsection{Proof of Theorem 1}
\label{Apx4}

\begin{theorem}
\label{th1}
For a state-action pair $(s_t, a_t)$, we define $\epsilon_r(s_t, a_t)$ as the reward gap between the reward given by the real environment and the learned dynamics model. And $\epsilon_{r_{max}}$ represents the maximum reward gap of all state-action pairs defined on $\mathcal{S}$ and $\mathcal{A}$. Denote the model error by $\epsilon_m = max_t[D_{TV}(p(s_{t+1}|s_t,a_t)||\hat{p}(s_{t+1}|s_t,a_t))]$. Given the trajectory $\delta(s_0, a_0)$ starting from $(s_0, a_0)$, the TREE error can be bounded as:
\begin{equation}\label{eq4_2}
\begin{split}
Error_{TREE} = J_{env}[\delta(s_0, a_0)]-J_{model}[\delta(s_0, a_0)] \leq \underbrace{\frac{(1-\gamma^{H})\epsilon_{r_{max}}}{1-\gamma}}_{\text{reward  estimation  gap}} + 
\underbrace{\frac{2r_{max}(\gamma-\gamma^{H})\epsilon_m}{1-\gamma}}_{\text{model  error}}
\end{split}
\end{equation}
\end{theorem}

\begin{proof}

Denote the predicted state distribution given by the learned dynamics model $\hat{f}$ as  $\hat{p}(s_{t+1}|s_t,a_t)$, the real state distribution given by the environment as $p(s_{t+1}|s_t,a_t)$ 
and the action distribution given by the policy as $p_{\pi}(a_t|s_t)$. We can prove Theorem 1 as follows:

\begin{equation*}
\begin{aligned}
&J_{env}[\delta(s_0, a_0)]-J_{model}[\delta(s_0, a_0)] \\
&=r_{e}(s_0,a_0)-r_{m}(s_0,a_0) + \sum^{H-1}_{t=1}\sum_{s_t,a_t}{\gamma^{t}[p(s_t,a_t)r_{e}(s_t,a_t)-\hat{p}(s_t,a_t)r_{m}(s_t,a_t)]} \\
&=r_{e}(s_0,a_0)-r_{m}(s_0,a_0) + \sum^{H-1}_{t=1}\sum_{s_t,a_t}{\gamma^{t}[p(s_t| s_{t-1},a_{t-1}) p_{\pi}(a_{t}|s_{t}) r_{e}(s_t,a_t)} \\&\quad \quad\quad \quad \quad \quad\quad \quad \quad \quad\quad \quad \quad \quad \quad \quad -{\hat{p}(s_t| s_{t-1},a_{t-1}) p_{\pi}(a_{t}|s_{t})  r_{m}(s_t,a_t)]} \\
&=r_{e}(s_0,a_0)-r_{m}(s_0,a_0) + \sum^{H-1}_{t=1}\sum_{s_t,a_t}{\gamma^{t}p_{\pi}(a_{t}|s_{t})[p(s_t| s_{t-1},a_{t-1}) r_{e}(s_t,a_t)} \\&\quad \quad\quad\quad \quad \quad \quad\quad \quad \quad \quad\quad \quad \quad \quad \quad -{\hat{p}(s_t| s_{t-1},a_{t-1})  r_{m}(s_t,a_t)]} \\
&= r_{e}(s_0,a_0)-r_{m}(s_0,a_0)+\sum^{H-1}_{t=1} \sum_{s_t,a_t}{\gamma^{t}p_{\pi}(a_{t}|s_{t})[p(s_t| s_{t-1},a_{t-1}) r_{e}(s_t,a_t)-\hat{p}(s_t| s_{t-1},a_{t-1})r_{e}(s_t,a_t)}\\& \quad \quad \quad \quad\quad \quad \quad \quad \quad \quad \quad \quad \quad \quad \quad \quad{+\hat{p}(s_t| s_{t-1},a_{t-1})r_{e}(s_t,a_t) -\hat{p}(s_t| s_{t-1},a_{t-1})r_{m}(s_t,a_t)]} \\
&=r_{e}(s_0,a_0)-r_{m}(s_0,a_0) + \sum^{H-1}_{t=1}\sum_{s_t,a_t}{\gamma^{t}p_{\pi}(a_{t}|s_{t})[[p(s_t| s_{t-1},a_{t-1})-\hat{p}(s_t| s_{t-1},a_{t-1})]r_{e}(s_t,a_t)}\\& \quad \quad \quad \quad \quad \quad \quad \quad\quad \quad \quad \quad \quad \quad\quad \quad{+\hat{p}(s_t| s_{t-1},a_{t-1})[r_{e}(s_t,a_t)-r_{m}(s_t,a_t)]]} \\
&=r_{e}(s_0,a_0)-r_{m}(s_0,a_0) + \sum^{H-1}_{t=1}\sum_{s_t,a_t}{\gamma^{t}p_{\pi}(a_{t}|s_{t})[p(s_t| s_{t-1},a_{t-1})-\hat{p}(s_t| s_{t-1},a_{t-1})]r_{e}(s_t,a_t)}\\& \quad \quad \quad \quad \quad \quad \quad \quad\quad \quad \quad  {+\sum^{H-1}_{t=1}\sum_{s_t,a_t}\gamma^{t}p_{\pi}(a_{t}|s_{t})\hat{p}(s_t| s_{t-1},a_{t-1})[r_{e}(s_t,a_t)-r_{m}(s_t,a_t)]} \\
&=r_{e}(s_0,a_0)-r_{m}(s_0,a_0) + \sum^{H-1}_{t=1}\sum_{s_t,a_t}{\gamma^{t}p_{\pi}(a_{t}|s_{t})[p(s_t| s_{t-1},a_{t-1})-\hat{p}(s_t| s_{t-1},a_{t-1})]r_{e}(s_t,a_t)}\\& \quad \quad \quad \quad \quad \quad \quad \quad\quad \quad \quad  {+\sum^{H-1}_{t=1}\gamma^{t}E_{s_{t}\sim \hat{p}, a_{t}\sim\pi}[r_{e}(s_t,a_t)-r_{m}(s_t,a_t)]} \\
\end{aligned}
\end{equation*}

\begin{equation*}
\begin{aligned}
&\leq \epsilon_{r_{max}}+ \sum^{H-1}_{t=1}\gamma^{t}r_{max} \sum_{s_t} |p(s_t| s_{t-1},a_{t-1})-\hat{p}(s_t| s_{t-1},a_{t-1})| + \sum^{H-1}_{t=1}\gamma^{t}\epsilon_{r_{max}} \quad \quad \quad \quad \quad \quad \quad \quad\quad \quad \quad \\
&= \epsilon_{r_{max}}+ \sum^{H-1}_{t=1}\gamma^{t}r_{max} 2D_{TV}(p(s_{t}|s_{t-1},a_{t-1})||\hat{p}(s_{t}|s_{t-1},a_{t-1})) + \sum^{H-1}_{t=1}\gamma^{t}\epsilon_{r_{max}}\\
&\leq \epsilon_{r_{max}}+ \sum^{H-1}_{t=1}{\gamma^{t}[2r_{max}\epsilon_{m}+\epsilon_{r_{max}}]} \\
&=\epsilon_{r_{max}}+ \frac{\gamma[2r_{max}\epsilon_{m}+\epsilon_{r_{max}}](1-\gamma^{H-1})}{1-\gamma} \\
&=\frac{(1-\gamma^{H})\epsilon_{r_{max}}}{1-\gamma} + \frac{2r_{max}(\gamma-\gamma^{H})\epsilon_m}{1-\gamma}
\end{aligned}
\end{equation*}
\end{proof}

\subsection{Continues Control Benchmark Tasks in Section 6}

\begin{figure}[h]
\centering
\subfigure[HalfCheetah]{              
\includegraphics[width=3cm,height=2.5cm]{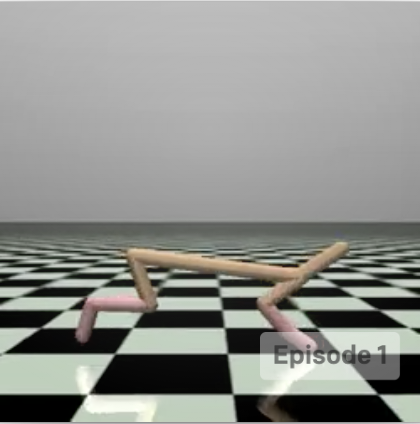}}
\hspace{0in}
\subfigure[Ant]{
\includegraphics[width=3cm,height=2.5cm]{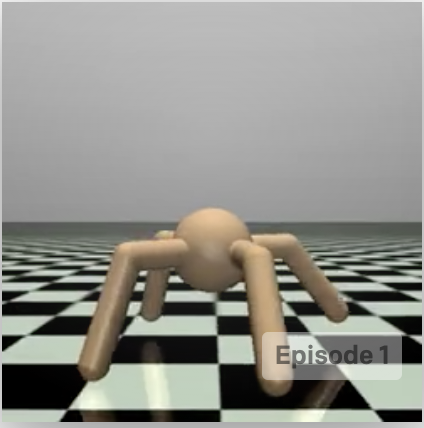}}
\hspace{0in}
\subfigure[Walker2D]{
\includegraphics[width=3cm,height=2.5cm]{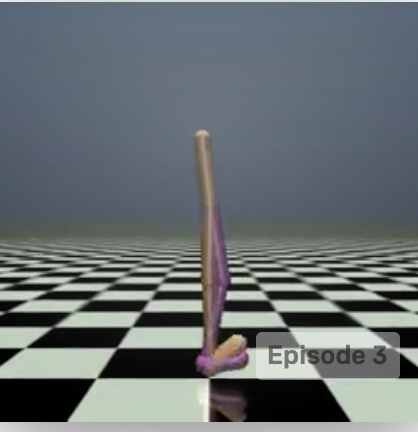}}
\hspace{0in}
\subfigure[Cheetah]{
\includegraphics[width=3cm,height=2.5cm]{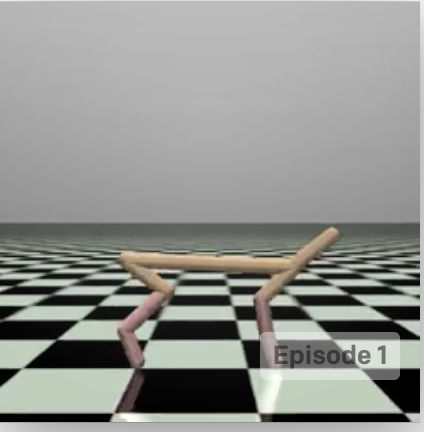}}
\caption{Visualization of four tasks.}
\label{fga1}
\end{figure}

In this section, we provide an overview of the continues control benchmark tasks used in Section. 6.  
Figure \ref{fga1} visualizes these four tasks and Table \ref{tba1} shows the dimensionality and horizon lengths of those tasks.

\begin{table}[h]
\setlength{\abovecaptionskip}{0.2cm}
\centering
\caption{Dimensions of observation and action space, and task horizon length for four tasks.}
\begin{tabular}{ccccc}
\toprule 
Environment name&Observation Space Dimension&Action Space Dimension&Task Horizon \\
\midrule
HalfCheetah& $18$ & $6$ & $1000$ \\
Ant& $28$ &$8$&$1000$ \\
Walker2D& $17$ &$6$&$1000$ \\
Cheetah& $17$ &$6$&$1000$ \\
\bottomrule
\end{tabular}
\label{tba1}
\end{table}

\clearpage

\subsection{Ablation Study}
\begin{figure}[h]
\centering
\subfigure[Ant]{              
\includegraphics[width=6cm,height=4.5cm]{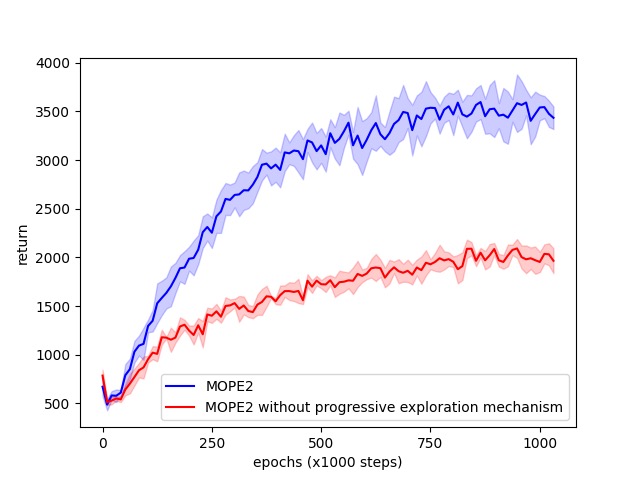}}
\hspace{0in}
\subfigure[Walker2D]{
\includegraphics[width=6cm,height=4.5cm]{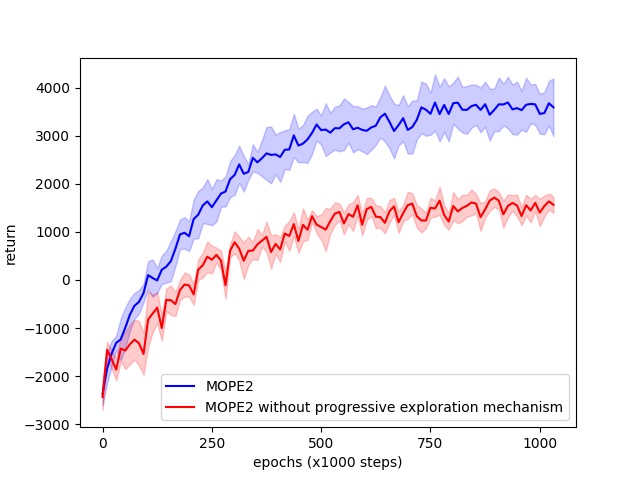}}
\caption{Ablation experiment results of MOPE2 and MOPE2 without progressive exploration mechanism on Ant and Walker2D.}
\label{fga2}
\end{figure}

In this section, we conduct ablation experiments on Ant-v2 and Walker2D-v2 respectively to prove the effectiveness of the progressive exploration mechanism.
The performance curves are shown in Figure \ref{fga2}.

It can be seen from Figure \ref{fga2} that progressive exploration mechanism makes a huge difference. 
When the progressive exploration mechanism is not used, the existence of intrinsic reward makes the agent continue to explore the environment and its performance is constantly rising. 
However, due to the inaccuracy of the model in the early stage of model learning, the intrinsic reward will lead to the wrong exploration, resulting in the sample efficiency and asymptotic performance of MOPE2 without progressive exploration mechanism are far worse than that of MOPE2 with progressive exploration mechanism.

\subsection{More Benchmark Experiments}

\begin{table}[h]
\centering
\caption{Performance of different algorithms on 9 continues control benchmark tasks. The results show the mean and standard deviation averaged over four random seeds and a window size of 5000 interaction steps.}
\resizebox{\textwidth}{42mm}{
\begin{tabular}{cccccc}
\toprule  
&Swimmer&Acrobot&Pendulum&Pusher&CartPole \\
\midrule  
MOPE2 & $38.9 \pm 11.0$ & $\mathbf{28.4 \pm 33.9} $ & $\mathbf{185.1 \pm 23.8} $ & $\mathbf{-56.7 \pm 18.3}$ &$\mathbf{200.2\pm0.6}$ \\
PETS & $17.9 \pm 20.6 $ & $-15.3\pm29.5$ & $134.8\pm114.4$ & $-91.6\pm23.1$&$199.3\pm1.2$ \\
SLBO &$\mathbf{41.6\pm18.4}$ & $-75.6\pm8.8$ & $173.5\pm2.5$ & $-58.4\pm23.3$&$78.0\pm166.6$ \\
ME-TRPO & $31,6\pm9.7$ & $-68.1\pm6.7$ & $177.3\pm1.9$ & $-98.5\pm12.6$&$160.1\pm69.1$ \\
GPS&$14.5\pm5.6$&$-193.3\pm11.7$&$162.7\pm7.6$&$-147.2\pm5.1$ & $14.4\pm18.6$ \\
PILCO & $-13.8\pm16.1$ & $-394.4\pm1.4$ & $-132.6\pm410.1$ & $-104.2\pm11.3$ & $-1.9\pm155.9$ \\
\cmidrule{1-6}
PPO&$38.0\pm1.5$&$-95.3\pm8.9$&$163.4\pm8.0$&$-227.9\pm8.3$&$86.5\pm7.8$ \\
TD3&$21.5\pm23.9$&$-64.3\pm6.9$&$161.4\pm14.4$&$-203\pm56.3$&$196.0\pm3.1$ \\
SAC &$23.0\pm17.3$ &$-52.9\pm2.0$ &$168.2\pm9.5$ &$-233.8\pm57.1$ &$199.4\pm0.4$  \\
\cmidrule{1-6}
Time-step&200000&200000&200000&50000&200000 \\
\cmidrule{1-6}
&Reacher3D&InvertedPendulum&Hopper&Reacher \\
\cmidrule{1-6}
MOPE2&$\mathbf{-20.7\pm10.9}$&$\mathbf{-0.1\pm0.3}$&$1396.2\pm372.2$ &$-7.3\pm2.1$ &$$ \\
PETS&$-42.8\pm29.1$&$-52.4\pm43.6$&$-615.2\pm804.9$ &$-13.2\pm5.2$ &$$ \\
SLBO& $-29.4\pm16.7$&$-240.4\pm7.2$&$1132.7\pm274.6$&$\mathbf{-4.1\pm0.1}$&$$ \\
ME-TRPO& $-38.9\pm6.1$&$-126.2\pm86.6$&$1272.5\pm500.9$&$13.4\pm0.2$&$$ \\
GPS& $-646.3\pm457.8$&$-74.6\pm97.8$&$-768.5\pm200.9$&$-19.8\pm0.9$&$$ \\
PILCO & $-62.5\pm27.9$ & $-194.5\pm0.8$ & $-1729.9\pm1611.1$ & $-13.2\pm5.9$ & $$ \\
\cmidrule{1-6}
PPO& $-174.5\pm13.6$&$-40.8\pm21.0$&$-103.8\pm1028.0$&$-17.2\pm0.9$&$$ \\
TD3& $-431.9\pm173.6$&$-224.5\pm0.4$&$\mathbf{2245.3\pm232.4}$&$-14.0\pm0.9$&$$ \\
SAC& $-147.3\pm34.8$&$-0.2\pm0.1$&$726.4\pm675.5$&$-6.4\pm0.5$&$$ \\
\cmidrule{1-6}
Time-step&50000&200000&200000&200000& \\
\bottomrule
\end{tabular}}
\label{tba2}
\end{table}

\begin{figure}[h]
\centering
\subfigure[Hopper]{              
\includegraphics[width=4cm,height=3cm]{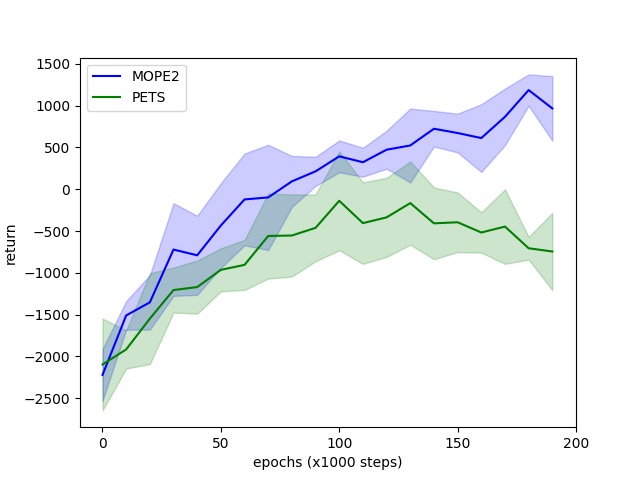}}
\hspace{0in}
\subfigure[Acrobot]{
\includegraphics[width=4cm,height=3cm]{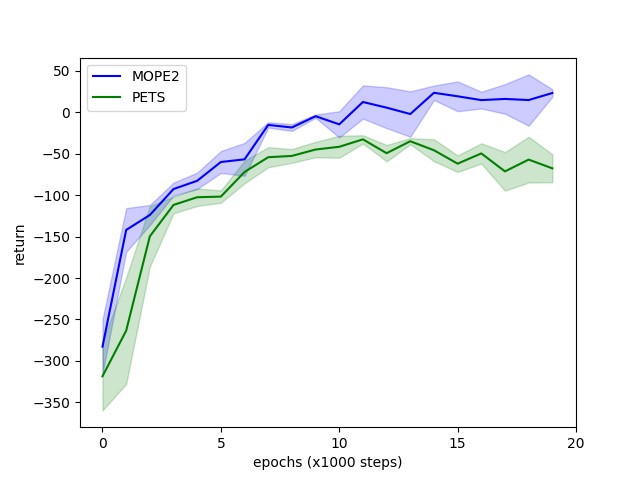}}
\hspace{0in}
\subfigure[Reacher]{
\includegraphics[width=4cm,height=3cm]{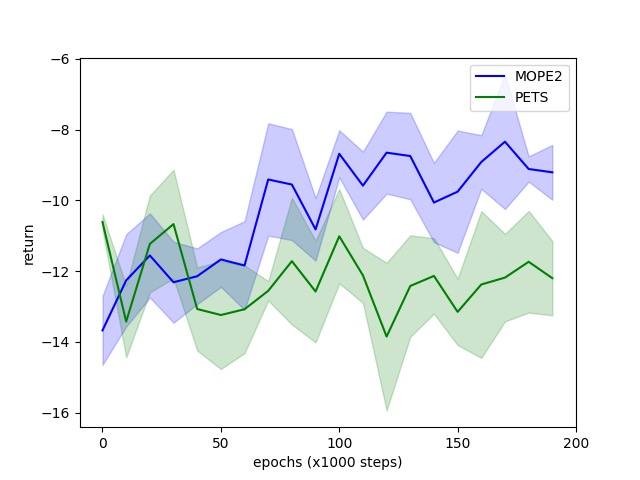}}
\hspace{0in}
\subfigure[Swimmer]{
\includegraphics[width=4cm,height=3cm]{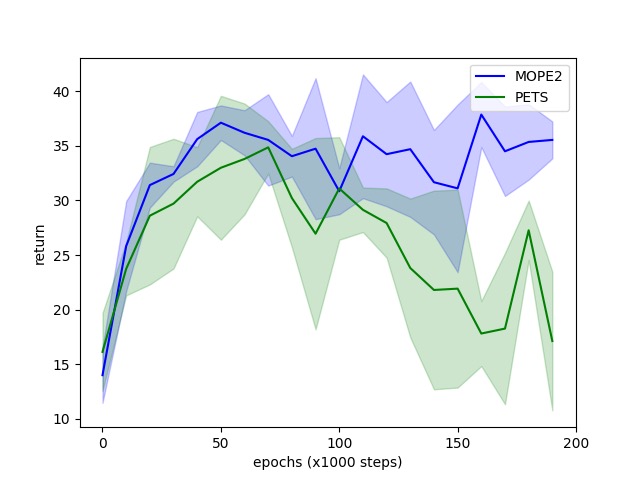}}
\hspace{0in}
\subfigure[CartPole]{
\includegraphics[width=4cm,height=3cm]{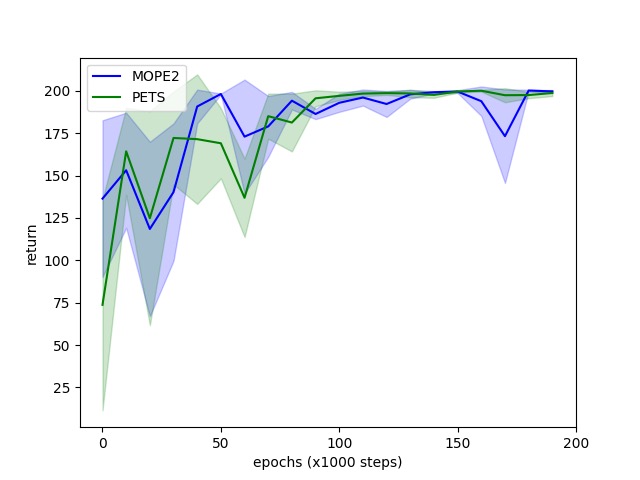}}`
\hspace{0in}
\subfigure[Pendulum]{
\includegraphics[width=4cm,height=3cm]{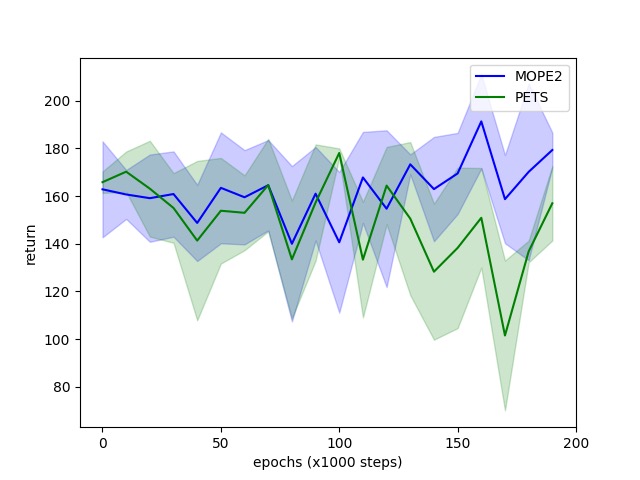}}
\hspace{0in}
\subfigure[InvertedPendulum]{
\includegraphics[width=4cm,height=3cm]{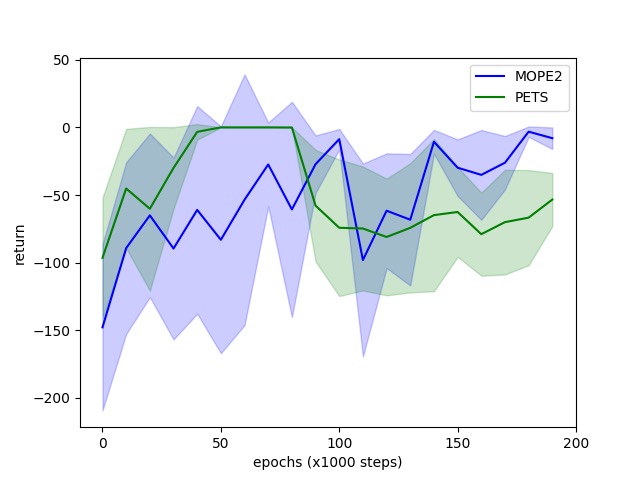}}
\hspace{0in}
\subfigure[Pusher]{
\includegraphics[width=4cm,height=3cm]{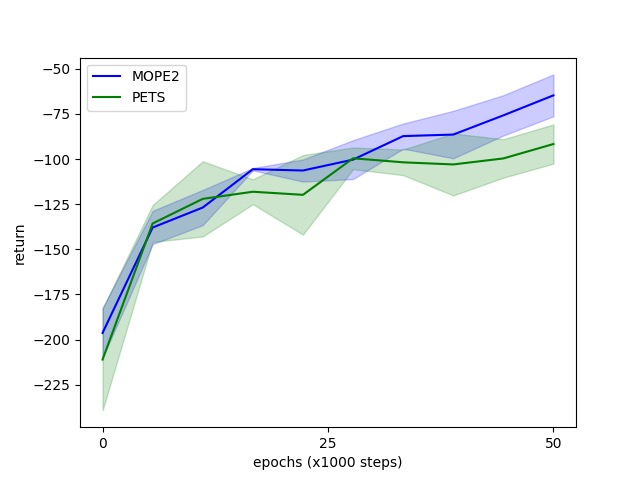}}
\hspace{0in}
\subfigure[Reacher3D]{
\includegraphics[width=4cm,height=3cm]{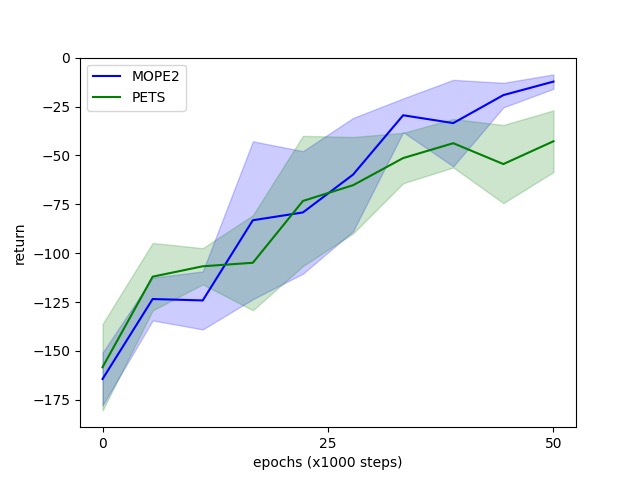}}
\caption{Performance curves of MOPE2 and PETS in 9 continues control benchmark tasks.}
\label{fga3}
\end{figure}

To prove the effectiveness and high sample efficiency of MOPE2, we conduct experiments with short interaction steps in another 9 continues control benchmark tasks.
In Pusher and Reacher3D, the number of interaction steps is set to 50K for these two tasks are relatively simple. And the number of interaction steps in the other 7 tasks is 200k.
The performance curves of MOPE2 and the state-of-the-art model-based control method PETS are shown in Figure \ref{fga3} and performance of different algorithms is summarized in Table \ref{tba2} \footnote{The performance of GPS, PILCO and PPO comes from the experiment results in Reference. 16.}.
From Table \ref{tba2} we can find that MOPE2 performs best in 6 tasks compared with both MBRL and MFRL methods. 
And MOPE2 achieve the best performance among MBRL methods in Hopper, just lags behind the MFRL method TD3. 
Moreover, in Swimmer and Reacher, MOPE2 is very close to the best performance.
On the other hand, by comparing the performance curves in Figure \ref{fga3}, we can see that the performance and sample efficiency of MOPE2 are far better than the state-of-the-art model-based control method PETS, which further proves the effectiveness of our method.

\subsection{Hyper-parameter Study}

In MOPE2, the most important parameter is $\beta_{max}$ which determines the upper limit of the agent's exploration ability. 
In this section, we conduct ablation experiments on Ant and Walker2D to verify the influence of different $\beta_{max}$ on the performance of MOPE2.
We set $\beta_{min}$, $e_{min}$ and $e_{max}$ as 0, 50, 300 respectively.

\begin{figure}[h]
\centering
\subfigure[Ant]{              
\includegraphics[width=6cm,height=4.5cm]{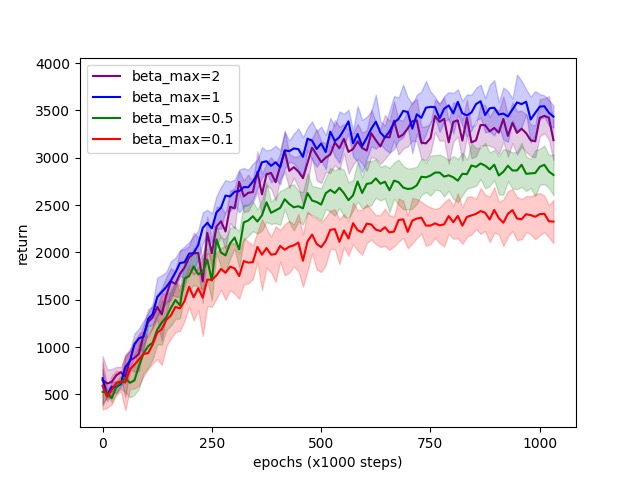}}
\hspace{0in}
\subfigure[Walker2D]{
\includegraphics[width=6cm,height=4.5cm]{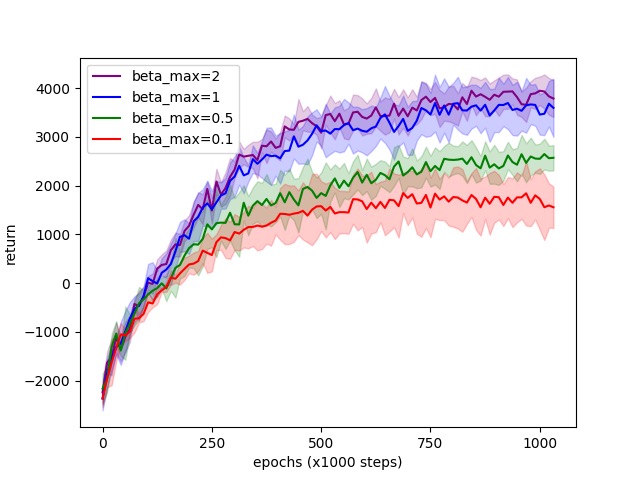}}
\caption{Ablation experiment results of MOPE2 and MOPE2 without progressive exploration mechanism on Ant and Walker2D.}
\label{fga4}
\end{figure}

The experiment results are shown in Figure. \ref{fga4}.
It can be seen that when $\beta_{max}$ is less than 1, the performance of MOPE2 is very sensitive to the choice of $\beta_{max}$. 
And the performance of MOPE2 continues to improve as $\beta_{max}$ increases. 
However, when $\beta_{max}$ exceeds 1, the choice of $\beta_{max}$ has little effect on the performance of MOPE2. 
MOPE2's performance on Ant even drops when $\beta_{max}$ equals 2. 
We think this is because the agent explores the out of the distribution actions and states. 
Therefore, in order to reasonably increase the agent's exploration ability, we set $\beta_{max}$ as 1 in all experiments in this paper.

\subsection{Hyper-parameters Settings}

\begin{table}[h]
\setlength{\abovecaptionskip}{0.2cm}
\centering
\caption{Hyper-parameters settings in our experiment.}
\begin{tabular}{ccc}
\toprule 
Hyper-parameter&Value \\
\midrule
$K$: generated trajectories' number & 500 \\
$H$: planning horizon & 30 \\
$k$: number of  chosen trajectories & 100 \\
$\alpha$: action distribution update rate & 0.01 \\
$\beta_{min}$ & 0 \\
$\beta_{max}$ & 1 \\
$e_{min}$ & 50 \\
$e_{max}$ & 300 \\
$e_{min}$ (in Pusher and Reacher3D) & 20 \\
$e_{max}$ (in Pusher and Reacher3D) & 50 \\
Initial Distribution Sigma & 0.1 \\
Initial Distribution Mu & 0 \\
Action distribution Iterations & 20 \\
Dynamics model ensemble size & 4 \\
Network architecture & MLP with 3 hidden layers of size 500 \\
Interaction steps per epoch & 1000 \\
\bottomrule
\end{tabular}
\label{tba3}
\end{table}

\end{document}